\pdfoutput=1

\documentclass[11pt]{article}

\usepackage[preprint]{acl}

\usepackage{times}
\usepackage{latexsym}

\usepackage[T1]{fontenc}

\usepackage[utf8]{inputenc}

\usepackage{microtype}

\usepackage{inconsolata}

\usepackage{graphicx}

\title{MedicalOS: An LLM Agent based Operating System for Digital Healthcare}

\author{Jared Zhu \\
  Independent Researcher \\
  \texttt{jaredzhu3@gmail.com} 
  \\\And
  Junde Wu \\
  University of Oxford \\ 
  \texttt{jundewu@ieee.org} \\
  }

\begin{document}
\maketitle
\begin{abstract}
Decades' advances in digital health technologies, such as electronic health records, have largely streamlined routine clinical processes. Yet, most these systems are still hard to learn and use: Clinicians often face the burden of managing multiple tools, repeating manual actions for each patient, navigating complicated UI trees to locate functions, and spending significant time on administration instead of caring for patients. The recent rise of large language model (LLM) based agents demonstrates exceptional capability in coding and computer operation, revealing the potential for humans to interact with operating systems and software not by direct manipulation, but by instructing agents through natural language. This shift highlights the need for an abstraction layer, an agent–computer interface, that translates human language into machine-executable commands. In digital healthcare, however, requires a more domain-specific abstractions that strictly follow trusted clinical guidelines and procedural standards to ensure safety, transparency, and compliance. To address this need, we present \textbf{MedicalOS}, a unified agent-based operational system designed as such a domain-specific abstract layer for healthcare. It translates human instructions into pre-defined digital healthcare commands, such as patient inquiry, history retrieval, exam management, report generation, referrals, treatment planning, that we wrapped as off-the-shelf tools using machine languages (e.g., Python, APIs, MCP, Linux). We empirically validate MedicalOS on 214 patient cases across 22 specialties, demonstrating high diagnostic accuracy and confidence, clinically sound examination requests, and consistent generation of structured reports and medication recommendations. These results highlight MedicalOS as a trustworthy and scalable foundation for advancing workflow automation in clinical practice.

\end{abstract}

\begin{figure*}[hbt!]
\vspace{-10pt}
    \centering
    \includegraphics[width=0.75\linewidth]{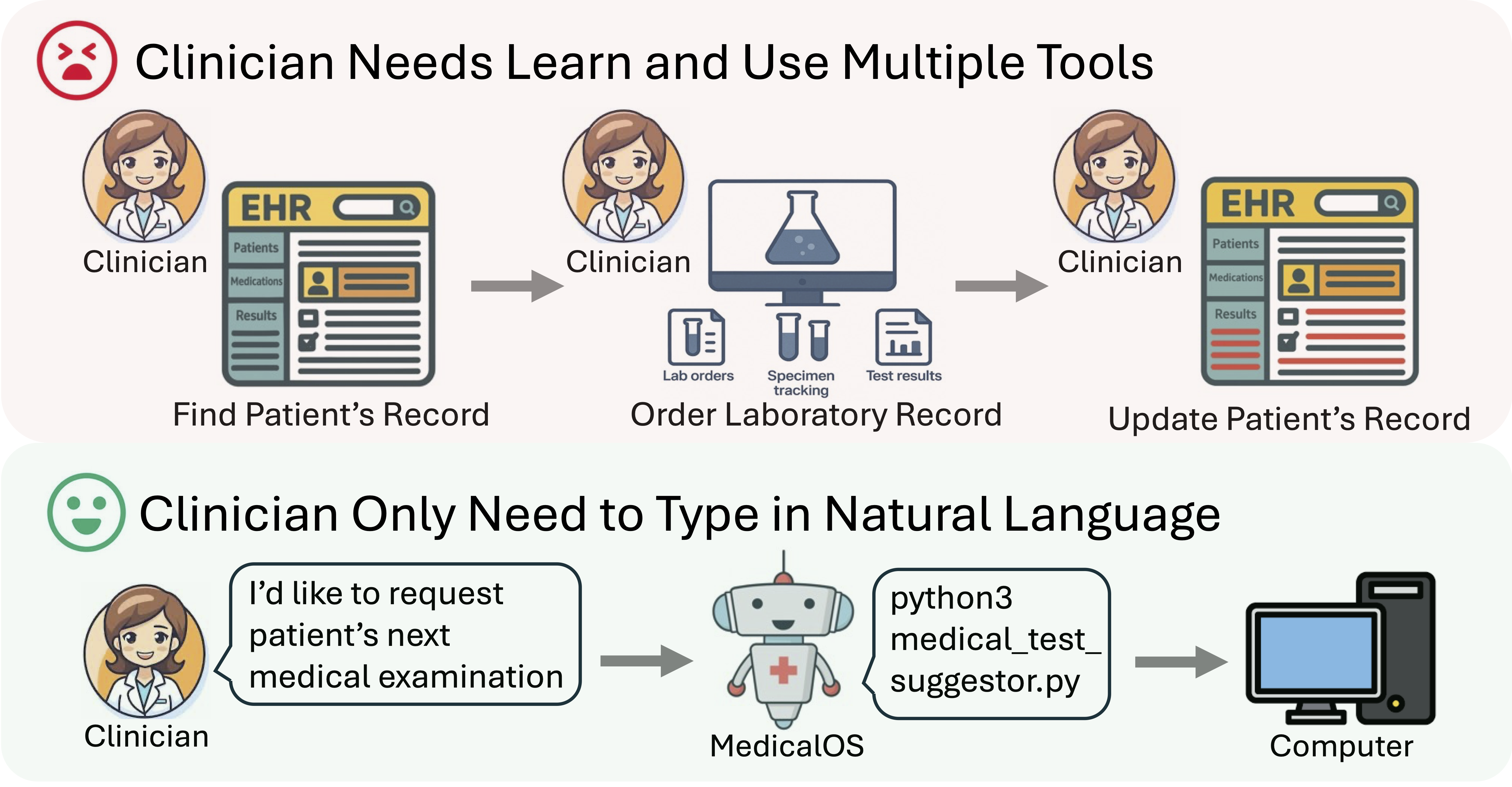} 
    \vspace{-5pt}
    \caption{MedicalOS Overview. }
    \label{fig:overview}
\vspace{-10pt}
\end{figure*}

\section{Introduction}
The digital transformation of healthcare, driven by the widespread adoption of cloud-based infrastructure and electronic health records (EHRs), has significantly reshaped clinical workflows \cite{menachemi2011benefits, king2014clinical, workneh2018understanding}. Clinicians now use digital systems to perform routine tasks such as patient registration, documentation, and referral, making these processes more systematic and accessible \cite{schopf2019well, barbieri2023electronic}. However, operating these systems still requires significant effort from healthcare professionals \cite{johnson2022making, hswen2023electronic}. Clinicians must invest substantial time in learning how to use various tools and continuously adapting to newly introduced features. This ongoing need for training increases cognitive load and reduces time available for direct patient care. In practice, even completing a single task often involves navigating multiple interfaces and performing a sequence of manual operations. These actions must be repeated for each patient encounter, compounding the administrative burden. Physicians now spend more than half of their working hours on EHRs and related documentation tasks \cite{bongurala2024transforming}. As a result, although current digital systems facilitate daily medical routines, they continue to place a considerable burden on clinicians. These systems remain labor-intensive, costly to scale, and difficult to full automation, leaving a gap between current digital systems and true workflow automation in clinical settings.

Recent advances in agent-based systems offer a promising path toward bridging the gap between human requirements expressed in natural language and machine execution based on formal commands \cite{yang2024swe, antoniades2024swe, yang2025swe}, enabling end-to-end automation. Traditionally, this gap has been mediated through graphical user interfaces (GUIs \cite{jansen1998graphical}), which require users to learn how to operate various tools by following instruction manuals and manually coordinating actions across systems. For example, planning a trip to Hawaii typically requires using platforms like Booking.com and Airbnb, selecting dates, comparing options, and finalizing reservations step by step. This manual burden is being addressed by systems such as command line interfaces \cite{schirano_doriandarkoclaude-engineer_2025}, which introduce an abstract layer between users and software. These systems interpret natural language input and translate it into structured commands that can be executed by machines. As a result, complex workflows like booking flights and selecting accommodations can be automated from beginning to end. These developments demonstrate that natural language can be directly used to drive machine-level execution through intermediate abstract layers, bypassing traditional GUIs and reducing the need for manual control. This paradigm introduces a promising opportunity in clinical settings: enabling clinicians to express medical goals in natural language while intelligent agents manage the underlying digital systems. Such an approach has the potential to reduce manual workload, improve scalability, and enable fully automated healthcare delivery.

However, applying this type of abstract layer to medicine introduces unique challenges. Unlike general-purpose command line systems that operate under flexible programming conventions, medical workflows require domain-specific abstract layers that strictly follow established clinical standards. Medicine is a highly structured and tightly regulated field, guided by established guidelines and procedural protocols \cite{ullah2024challenges, gencturk2024transforming, mennella2024ethical, kalokyri2025ai}. Authoritative resources such as \textit{Stedman's Medical Dictionary} \cite{stedman1920stedman}, the \textit{MSD Manual} \cite{noauthor_msd_nodate}, and the \textit{British National Formulary} (BNF \cite{noauthor_bnf_2025}) define controlled vocabulary, diagnostic references, and medication prescription that constrain how clinical tasks must be performed. At the same time, natural language is inherently ambiguous and often imprecise, whereas programming languages are designed to be precise and executable by computers. AI agents can serve as a powerful intermediary, translating high-level clinician requirements into structured symbolic actions. However, this translation process often involves inferring user intent and making default assumptions. While such a ``first-make-then-fix'' strategy is generally acceptable in domains like software engineering, it introduces substantial risk in high-stakes fields such as medicine. In clinical contexts, every step must be transparent, verifiable, and compliant with professional guidelines. Thus, translating a clinician’s instruction into machine-executable actions is analogous to compiling human intent into a domain-specific ``medical programming language''—one that 
must follow trusted clinical guidelines and established procedural routines. For automation to be feasible and trustworthy in healthcare, this abstract layer must not only understand natural language medical requirements but also translate them into precise and verifiable commands that can be processed by computers in accordance with established medical standards.

\begin{figure*}[hbt!]
\vspace{-10pt}
    \centering
    \includegraphics[width=0.98\linewidth]{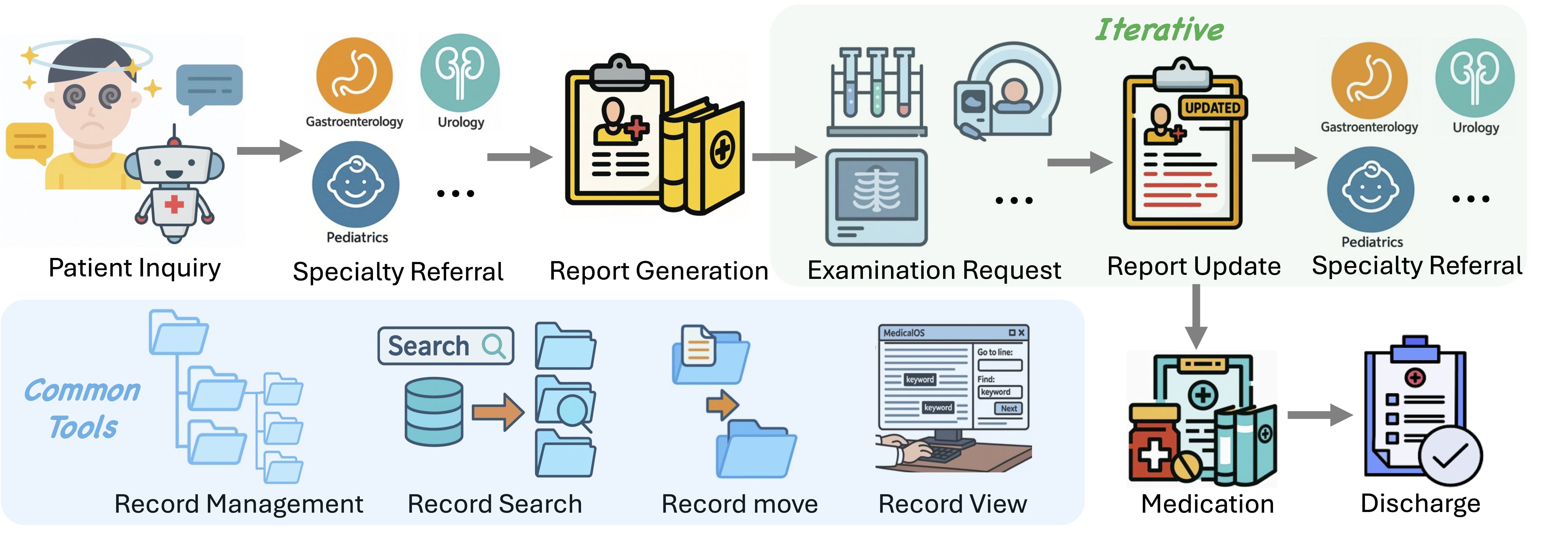} 
    \vspace{-5pt}
    \caption{MedicalOS Workflow. }
    \label{fig:workflow}
\vspace{-10pt}
\end{figure*}

To address these challenges, we propose \textbf{MedicalOS}, an agent-based \textbf{Medical Operational System} that functions as an abstract layer between high-level clinical requirements and low-level executable computer commands. Designed specifically for healthcare, MedicalOS is comprehensive enough to fulfill real-world clinical tasks while remaining strictly aligned with trusted medical guidelines. At its core, MedicalOS enables translating natural language instructions into executable commands through a reasoning-and-acting framework (ReAct \cite{yao2023react, wu2025agentic, singh2025agentic}) that operates under medical protocols. This framework interprets clinician requirements and determines when and how a series of medically valid actions should be executed, ensuring adherence to established clinical standards. MedicalOS supports a continuous and interpretable care pathway, encompassing patient inquiry, history retrieval, report generation, specialist referral, and treatment planning. 
As a result, MedicalOS offers a low-effort, scalable and trustworthy foundation for fully automated healthcare delivery. In summary, our contributions are:

\begin{itemize}
    \item We introduce MedicalOS, \textbf{a unified agent-based operational system} that bridges the gap between high-level clinician requirements expressed in natural language and low-level machine execution through domain-specific abstract layer tailored to healthcare.

    \item MedicalOS implements a ReAct-based framework that follows trusted clinical guidelines and established medical standards, \textbf{ensuring reliability and trustworthiness}.
    
   \item MedicalOS supports \textbf{end-to-end clinical workflow automation}, fulfilling real-world clinical requirements across patient inquiry, history retrieval, documentation, report generation, referral, and treatment planning.

   \item MedicalOS is \textbf{empirically validated on 214 patient cases across 22 specialties}, demonstrating high diagnostic accuracy and confidence, appropriate examination requests, and consistent generation of structured reports and medication recommendations.
\end{itemize}

\section{Methodology}
We describe how \textbf{MedicalOS} operates as an agent–based operational system that enables large language models (LLMs) to autonomously manage end-to-end clinical workflows within a structured and medically grounded environments. MedicalOS embeds AI agents in a domain-specific abstract layer, allowing them to interact with clincial systems in a way that is aligned with trusted medical guidelines and procedural standards. Agents in MedicalOS perform a range of tasks including patient inquiry dialogues, retrieving historical medical records, and identify relevant past cases. The system supports specialty referral with justifiable explanations, manages clinical documentations, and generates structured medical reports grounded in authoritative knowledge sources. Additionally, MedicalOS supports real-time report updates in response to newly available data, such as physical examination findings or laboratory test results, and clearly explains the reasoning behind each update. Medication recommendations are also produced using trusted medical knowledge bases, ensuring alignment with prescription standards. To support clinician-in-the-loop workflows, MedicalOS provides intuitive tools for document navigation and keyword search, facilitating low-effort yet reliable human verification. Agent actions are executed through a ReAct-based framework, enabling reasoning and acting cycles that iteratively refine behavior based on environmental feedback. Each step in this automatic medical workflow is transparent and trustworthy, ensuring that all operations remain compliant with professional clinical practice.

\subsection{Patient Inquiry} 
At the initial stage of interaction, MedicalOS functions as a virtual primary care physician powered by a LLM, engaging in natural language conversation with the patient to collect and document relevant clinical information. The entire dialogue is automatically transcribed and stored for future reference. Based on the patient's responses, MedicalOS evaluates whether further examination is needed. If so, it identifies the appropriate medical specialty for referral. If not, it provides medically grounded suggestions regarding potential medications and appropriate next steps. This process closely mirrors the clinical reasoning and triage decisions made by human primary care providers in real-world practice.

\subsection{Documentation Management} 
To support effective management of patient records, MedicalOS facilitates the retrieval of previous patient data through directory or document name search. This functionality helps clinicians quickly identify whether relevant historical records exist. If no prior patient data is found, MedicalOS automatically creates a new patient profile within the system. Once the appropriate directory is located, the transcribed conversation is moved to the corresponding patient folder. For cases requiring specialist referral, the corresponding specialty folder is first identified and the patient’s entire directory is moved into it. It enables the specialists to access and review the patient case within full context. These specialty folders are predefined with names used in the hospital, such as Cardiology or Oncology. In rare situations where a new specialty is introduced, a new folder can also be created. Each specialty folder contains records of the currently active patients under that specialty. MedicalOS also supports keyword-based search across all documents, returning a list of documents that contain the specified term. This enables clinicians to efficiently locate similar cases to inform current clinical decisions. For each matching document, the system can further highlight the exact lines where the keyword appears, facilitating rapid navigation an efficient access to relevant clinical information. To prevent information overload when there are too many matched cases, the search functionality includes support for truncated results, returning only the top-ranked matches by default.

\subsection{Report Generation}\label{sec:report generation}
When specialists review patient records, they are typically responsible for generating medical reports that summarize the patient's condition and provide recommendations for the next steps, \textit{e.g.,} requesting a chest CT scan. To support reliability and clinical grounding, MedicalOS first extracts up to three key terms from the initial patient inquiry conversation. These keywords are used to retrieve relevant sections from trusted medical sources, which serve as grounding information for report generation. In this work, we incorporate Wikipedia and PubMed as external knowledge bases, given their accessibility and coverage. Wikipedia provides broad, human-readable explanations of medical concepts, while PubMed offers authoritative, peer-reviewed clinical literature. The final report is generated in a structured format aligned with current clinical guidelines and includes seven core sections: patient identification, medical history, physical examination findings, test results, treatment plan, progress notes, and discharge summary \cite{bali2011management}. It also contains the sources of the grounded external knowledge. Once generated, the report is automatically stored within the patient's designated folder for future access and reference. Additionally, MedicalOS offers flexibility by supporting report generation based on a variety of inputs, including individual files, the complete patient folder, or a combination of both. This design accommodates different documentation styles and enables comprehensive summaries tailored to each clinical case. 

\subsection{Report Viewer}
Once a report is generated, clinicians are typically responsible for reviewing and verifying its contents before it is finalized or delivered to the patient. To support this process, MedicalOS provides an interactive report viewer designed for efficient navigation and verification. When dealing with lengthy documents, the system allows users to scroll through the report, jump directly to a specific line, and locate all instances of a given keyword. It also enables navigation to the first occurrence of the keyword within the text. These features streamline the review process, helping clinicians quickly assess critical content, verify factual accuracy, and make informed edits as needed. 

\subsection{Examination Request} \label{sec:Examination Request}
Beyond the initial assessment, clinicians may request additional examinations to support diagnosis, which requires updating medical reports as new information becomes available. MedicalOS analyzes the current patient report to determine whether further tests are needed. If so, the system automatically requests the appropriate examinations, such as lung biopsy or chest CT.

\subsection{Report Update} \label{sec:Report Update}
A revised report is then generated by integrating the newly acquired data and original report.
The updated report maintains the same structured seven-section format as mentioned in Section \ref{sec:report generation}. In addition, MedicalOS generates a supplementary explanation file that outlines and justifies any updates. If the diagnosis or treatment plan has changed, the system highlights the new information that caused the changes and explains the underlying reasoning behind it. This file enhances transparency and enables clinicians to easily verify modifications. Both the updated report and the corresponding explanation file are automatically stored in the patient's folder for future reference.

\subsection{Specialty Referral}\label{sec:Specialty Referral}
Following additional examinations, MedicalOS reevaluates the appropriate clinical specialty to determine whether a referral is necessary. Rather than simply outputting the target specialty, the system generates a structured referral report in accordance with established clinical conventions. This report includes the current managing specialty, the recommended specialty for referral, the rationale supporting the referral decision, and a brief summary of the patient's clinical status. It also contains a dedicated section titled ``Points for Attention," which highlights critical issues that the receiving specialist should consider to ensure continuity of care and clinical clarity. The referral report is stored in the patient's folder. Based on the referral outcome, MedicalOS then transfers the entire folder, including all relevant documentation, from the current specialty directory to the appropriate directory under the referred specialty.

\subsection{Medication}
MedicalOS repeats Report Update (Section \ref{sec:Report Update}) and Specialty Referral (Section \ref{sec:Specialty Referral}) until it gathers enough information, such as patient symptoms, examination findings, or laboratory results, to make a final diagnosis. It automatically decides whether the current information is sufficient. At the end of the clinical workflow, patients may be discharged with prescribed medications based on their final diagnosis. To ensure safety and adherence to clinical standards, MedicalOS generates medication recommendations grounded in trusted medical knowledge sources. In this work, we incorporate DailyMed and Wikipedia as reference materials. DailyMed, a database managed by the U.S. National Library of Medicine under the National Institutes of Health, is widely regarded as a reliable source for evidence-based prescribing practices, while Wikipedia offers accessible summaries of drug mechanisms and clinical considerations. Based on the most likely diagnosis identified in the previously generated medical report, MedicalOS determines appropriate treatment options by referencing DailyMed or Wikipedia. 
The final output includes detailed medication information such as brand and generic names, dosage, frequency, duration, key cautions, potential side effects, and patient-specific considerations. All external sources accessed during this process are documented in the output to support transparency and facilitate clinician verification. 

\subsection{Discharge}
Once treatment is completed, the patient is discharged, marking the end of their current care episode. At this point, MedicalOS automatically transfers the patient’s folder from the current specialty unit back to the central patient database. If the patient returns for future care, the system follows the same procedure, and all previously recorded information can be easily accessed and reviewed. This ensures continuity of care and allows clinicians to build upon prior clinical history without repeating earlier steps.

\begin{table*}[]
\centering
\caption{Diagnosis Accuracy}
\resizebox{1\linewidth}{!}{
\begin{tabular}{c|ccccccccccc|c}
\hline
\hline
Setting                                                               & Dermatology      & Psychiatry       & Gastroenterology & Neurology        & Cardiology       & Hematology       & Infectious Disease & Oncology         & Orthopedics      & Nephrology       & Pulmonology      & Overall          \\ \hline
CLI                                                                   & 89.43\%          & 83.28\%          & 85.55\%          & 86.58\%          & 82.61\%          & 78.04\%          & 84.21\%            & 79.83\%          & 86.01\%          & 79.19\%          & 89.79\%          & 84.70\%          \\
\begin{tabular}[c]{@{}c@{}}MedicalOS w/o \\ Test Request\end{tabular} & 89.31\%          & 84.07\%          & 86.98\%          & 86.18\%          & 82.18\%          & 79.33\%          & 84.74\%            & 79.50\%          & 86.44\%          & 80.62\%          & 91.92\%          & 84.98\%          \\
\begin{tabular}[c]{@{}c@{}}MedicalOS w/ \\ Test Request\end{tabular}  & \textbf{94.47\%} & \textbf{85.96\%} & \textbf{92.22\%} & \textbf{94.43\%} & \textbf{94.33\%} & \textbf{83.80\%} & \textbf{87.70\%}   & \textbf{86.97\%} & \textbf{93.25\%} & \textbf{88.30\%} & \textbf{95.43\%} & \textbf{90.24\%} \\ \hline \hline
\end{tabular}
}\label{tab:diagnosis accuracy}
\end{table*}

\begin{table*}[]
\centering
\caption{Diagnosis Confidence}
\resizebox{1\linewidth}{!}{
\begin{tabular}{c|ccccccccccc|c}
\hline
\hline
Setting                                                               & Dermatology   & Psychiatry    & Gastroenterology & Neurology     & Cardiology    & Hematology    & Infectious Disease & Oncology             & Orthopedics   & Nephrology    & Pulmonology   & Overall       \\ \hline
CLI                                                                   & 6.44          & 6.45          & 6.38             & 5.94          & 5.71          & 5.58          & 6.00               & 5.72                 & 6.54          & 5.72          & 6.70          & 6.21          \\
\begin{tabular}[c]{@{}c@{}}MedicalOS w/o \\ Test Request\end{tabular} & 5.68          & 5.72          & 5.67             & 5.29          & 5.07          & 5.17          & 5.25               & 5.09                 & 5.64          & 5.36          & 5.80          & 5.50          \\
\begin{tabular}[c]{@{}c@{}}MedicalOS w/ \\ Test Request\end{tabular}  & \textbf{7.76} & \textbf{6.73} & \textbf{7.44}    & \textbf{7.41} & \textbf{7.14} & \textbf{6.33} & \textbf{7.00}      & \textbf{7.55} & \textbf{7.45} & \textbf{7.00} & \textbf{7.60} & \textbf{7.19} \\ \hline \hline
\end{tabular}
}\label{tab:diagnosis confidence}
\end{table*}

\section{Experiment}

\subsection{Dataset}
To evaluate our proposed MedicalOS, we use the AgentClinic-MedQA dataset \cite{schmidgall2024agentclinic}. This dataset includes patient actor profiles with information such as demographics and reported symptoms, which we use to simulate patient responses during the inquiry process. For each case, we also extract past medical history from the patient actor profile and store it separately in the patient's folder, emulating how hospitals maintain historical medical records. The dataset further includes physical examination findings (\textit{e.g.,} vital signs) and test results (\textit{e.g.,} electrocardiograms), which are only accessible to MedicalOS upon explicit request. On average, each patient has 2.54 physical examination findings and 2.49 test results. Final evaluations are based on the ground truth diagnosis provided for each case. In total, the dataset comprises 214 cases across 22 specialties, with the most common being Dermatology (11.68\%), Psychiatry (10.28\%), and Gastroenterology (8.41\%).

\subsection{Experimental Setup} \label{sec:setup}
We define two main directories to support the experiments: \textit{Patient}, which contains folders for all patients along with their historical medical records, and \textit{Specialty}, which includes 22 predefined medical specialties. During the interaction process, patient records may move between these directories to reflect scenarios such as admission, referral, or discharge. To manage the diagnostic process, we set a maximum limit of four additional examinations that MedicalOS can request. This constraint ensures that, for evaluation purposes, the system is required to provide a final diagnosis even if uncertainty remains. If a requested examination or test is not available in the dataset, MedicalOS proceeds to the next inquiry cycle without receiving that specific information.

\subsection{Evaluation Matrix}
We employ OpenAIEmbeddings \cite{noauthor_openai_nodate} to generate numerical embeddings for both the predicted and ground truth diagnoses, thereby capturing their semantic relationships. Cosine similarity is then calculated between each pair to quantify alignment, and the resulting similarity score is used as a measure of diagnostic accuracy. Alongside the predicted diagnosis, MedicalOS also outputs a confidence score on a scale of 1 to 10, where lower values (1–4) indicate vague or uncertain findings with multiple possible causes, mid-range values (5–6) reflect moderate confidence supported by partial evidence, and higher values (7–10) represent strong or definitive diagnostic support. A final diagnosis is accepted, and no further examinations are required, when the confidence score exceeds 7 or meet the criteria in Section \ref{sec:setup}.

\subsection{Experimental Results}
\subsubsection{Diagnosis Accuracy \& Confidence}
We evaluate three different settings: (1) \textbf{CLI}, where the model generates diagnoses solely based on the patient-clinician conversation without any external information; (2) \textbf{MedicalOS w/o Test Request}, where the model has access to external medical resources but cannot request additional tests; and (3) \textbf{MedicalOS w/ Test Request}, the full system that combines external knowledge with the capability to suggest and incorporate further diagnostic tests.  

As shown in Table~\ref{tab:diagnosis accuracy}, MedicalOS w/ Test Request achieves the highest overall diagnosis accuracy at 90.24\%, outperforming both CLI (84.70\%) and MedicalOS w/o Test Request (84.98\%). The performance gains are particularly notable in Pulmonology (95.43\%), Orthopedics (93.25\%), and Neurology (94.43\%), where additional diagnostic testing appears to play a significant role in improving decision-making. Even in specialties with relatively high baseline accuracy, such as Dermatology and Gastroenterology, the full MedicalOS pipeline consistently yields improvements.

Table~\ref{tab:diagnosis confidence} presents the corresponding confidence scores. The full MedicalOS pipeline yields the highest overall confidence (7.19), exceeding the clinically acceptable threshold of 7. In contrast, MedicalOS w/o Test Request produces lower confidence (5.50) than the CLI setting (6.21), despite achieving slightly higher accuracy (84.98\% vs. 84.70\%). This suggests that while access to medical knowledge can improve prediction alignment, the ability to request further tests is critical for making confident, actionable decisions. The largest confidence improvements are observed in specialties where diagnosis is complex, such as Gastroenterology, Neurology, and Pulmonology.

Together, these results demonstrate that the integration of structured external resources and test-driven interaction significantly enhances both the accuracy and reliability of AI-assisted diagnosis across a wide range of clinical domains.

\subsubsection{Specialty Referral Analysis}
Among the 214 patient cases, 48 required at least one specialty referral, and two cases involved two separate specialty referrals. In two instances, MedicalOS failed to generate a referral report, despite clear indications that one was needed. At the first referral attempt, MedicalOS correctly identified the target specialty in 50\% of the cases when compared with the ground truth. After incorporating test results and follow-up interactions, this accuracy increased to 62.15\%. Overall, the results suggest that while the initial referral decisions show moderate alignment with clinical expectations, the iterative reasoning and test-request capabilities of MedicalOS substantially improve referral precision.

\subsubsection{Medical Examination Analysis}
\begin{figure}[ht]
            \centering
            \vspace{-5pt}\includegraphics[scale=0.3]{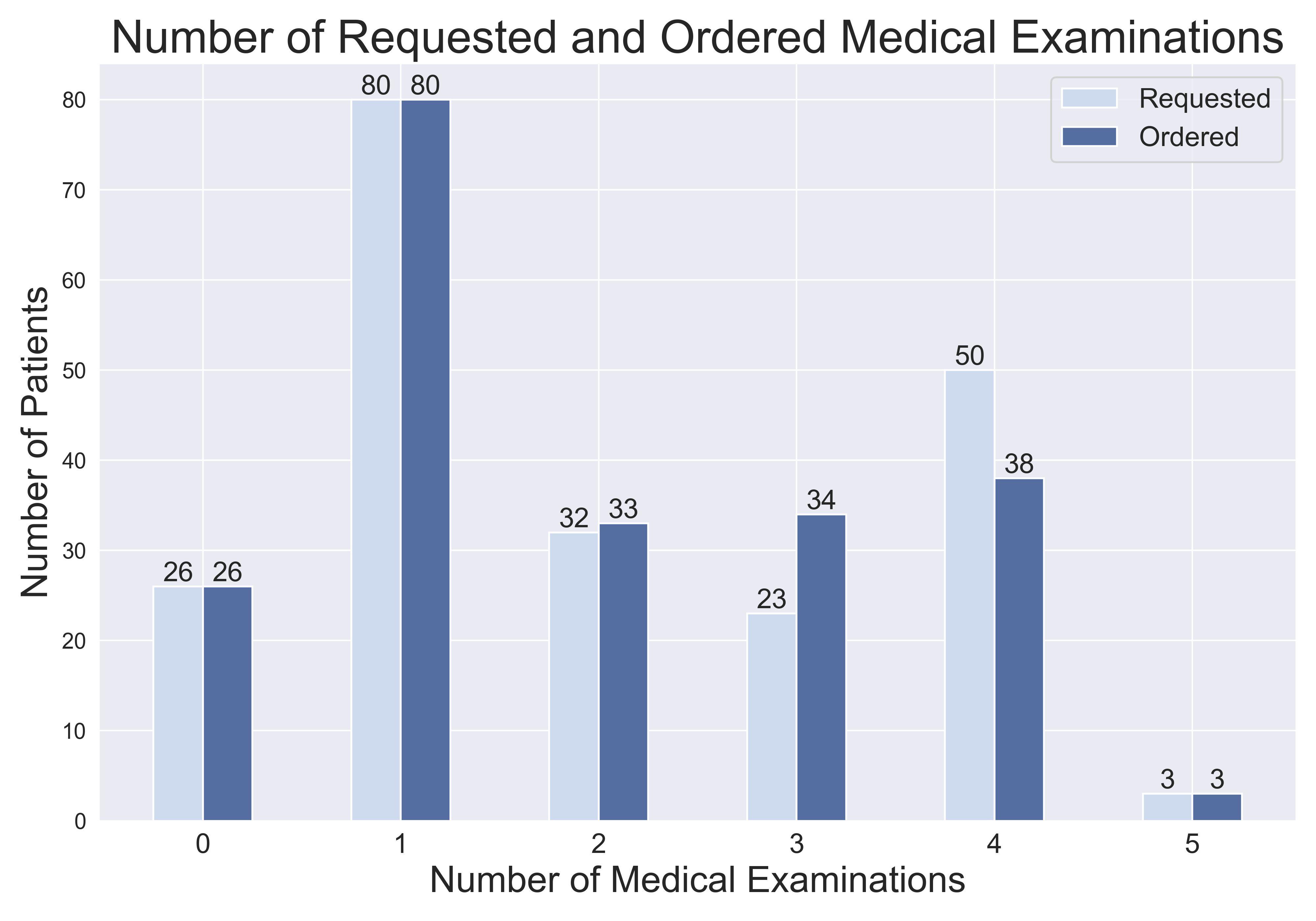}
            \vspace{-20pt}
    \caption{\textbf{Medical Examination Counts.}}
            \label{fig:exam_counts}
\vspace{-5pt}
\end{figure}
Figure~\ref{fig:exam_counts} illustrates the distribution of requested and ordered medical examinations across the 214 cases. In 26 cases, the patient inquiry conversation itself provided sufficient information for a confident diagnosis, and no additional examinations were required. The majority of patients, however, needed further testing: 37.38\% of cases reached a final diagnosis after a single examination, while smaller proportions required two (32 cases), three (23 cases), or four examinations (50 cases). Only three patients required as many as five examinations. The gap between requested and ordered examinations highlights an important limitation. While MedicalOS can identify the need for further testing, it does not always successfully follow through in ordering those tests. This discrepancy becomes more apparent as the number of examinations increases, suggesting that performance may be affected in more complex diagnostic scenarios. Nevertheless, the overall trend demonstrates that MedicalOS is capable of reducing unnecessary testing by reaching confident diagnoses in many cases after only one or two examinations.

\subsubsection{Ordered Medical Examination Analysis}
\begin{figure}[ht]
            \centering
            \vspace{-5pt}\includegraphics[scale=0.22]{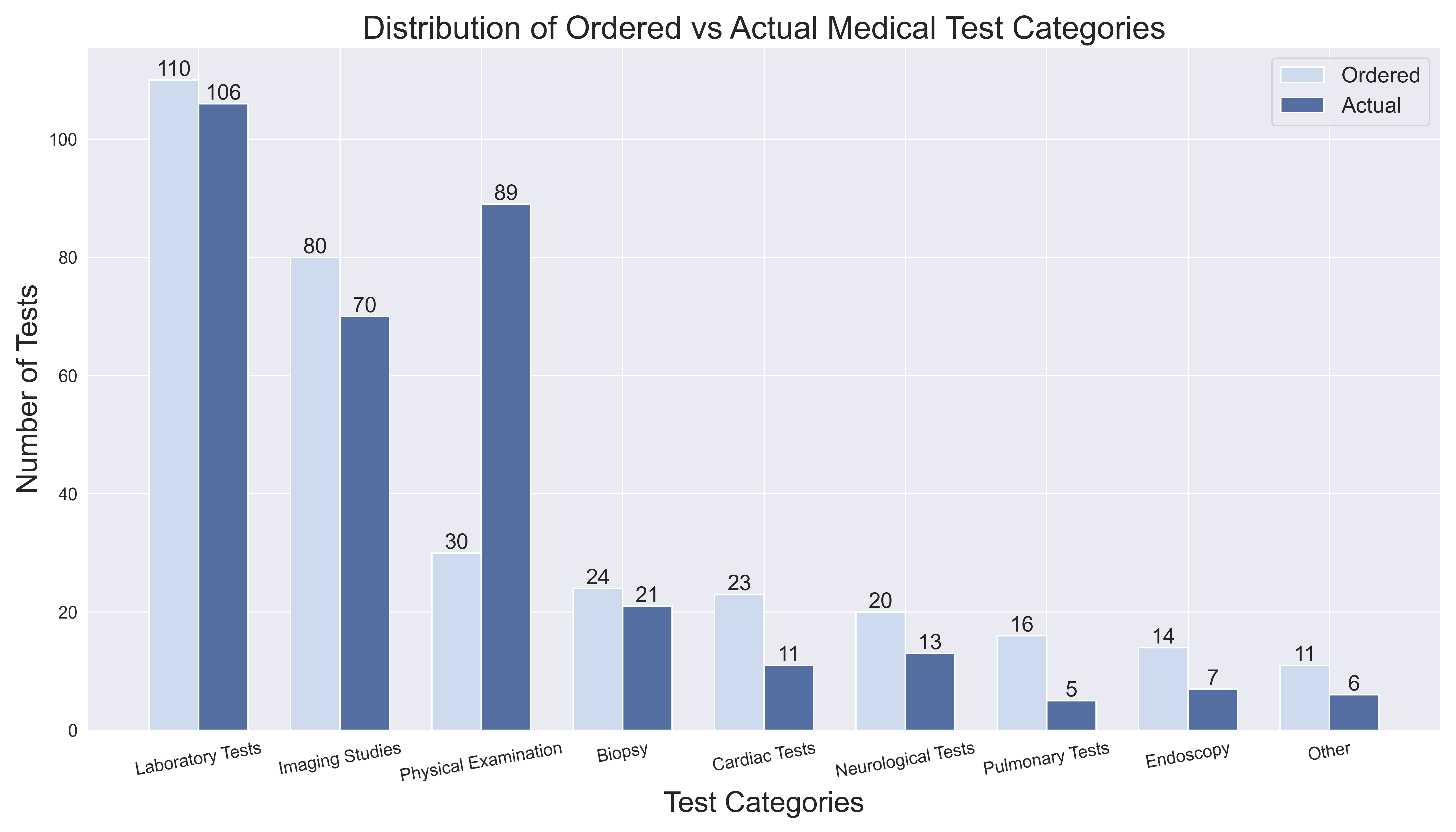}
            \vspace{-10pt}
    \caption{\textbf{Ordered Medical Examination Category.}}
            \label{fig:exam_category}
\vspace{-5pt}
\end{figure}
When a test is requested, MedicalOS maps it to the most relevant available examination in the patient record. For instance, if an abdominal ultrasound is requested but not directly available, the system retrieves the closest substitute, such as an abdominal examination report, from the dataset. Across all cases, a total of 415 examinations were ordered. Among these, 87 requests could not be matched with any similar or relevant results in the AgentClinic-MedQA dataset and were therefore skipped for the next iteration.  

Figure~\ref{fig:exam_category} presents the distribution of ordered tests grouped into major categories, compared with the actual examinations contained in the dataset. Laboratory tests form the largest category, with 135 ordered and 105 actual results, followed by imaging studies (80 ordered, 70 actual) and physical examinations (88 ordered, 80 actual). Overall, the distribution indicates that MedicalOS aligns closely with the examinations available in the dataset, reflecting its capability in requesting tests that are consistent with recorded clinical practice. Some discrepancies remain, particularly when requested tests do not have direct equivalents in the dataset. From a clinical perspective, the system’s strong reliance on laboratory, imaging, and physical examinations reflects common diagnostic workflows in real practice, where these modalities often provide the most direct evidence for decision-making. This alignment suggests that MedicalOS is capable of reproducing clinically meaningful diagnostic strategies.

\subsubsection{Medical Report Analysis}
\begin{figure}[ht]
            \centering
            \vspace{-10pt}\includegraphics[scale=0.05]{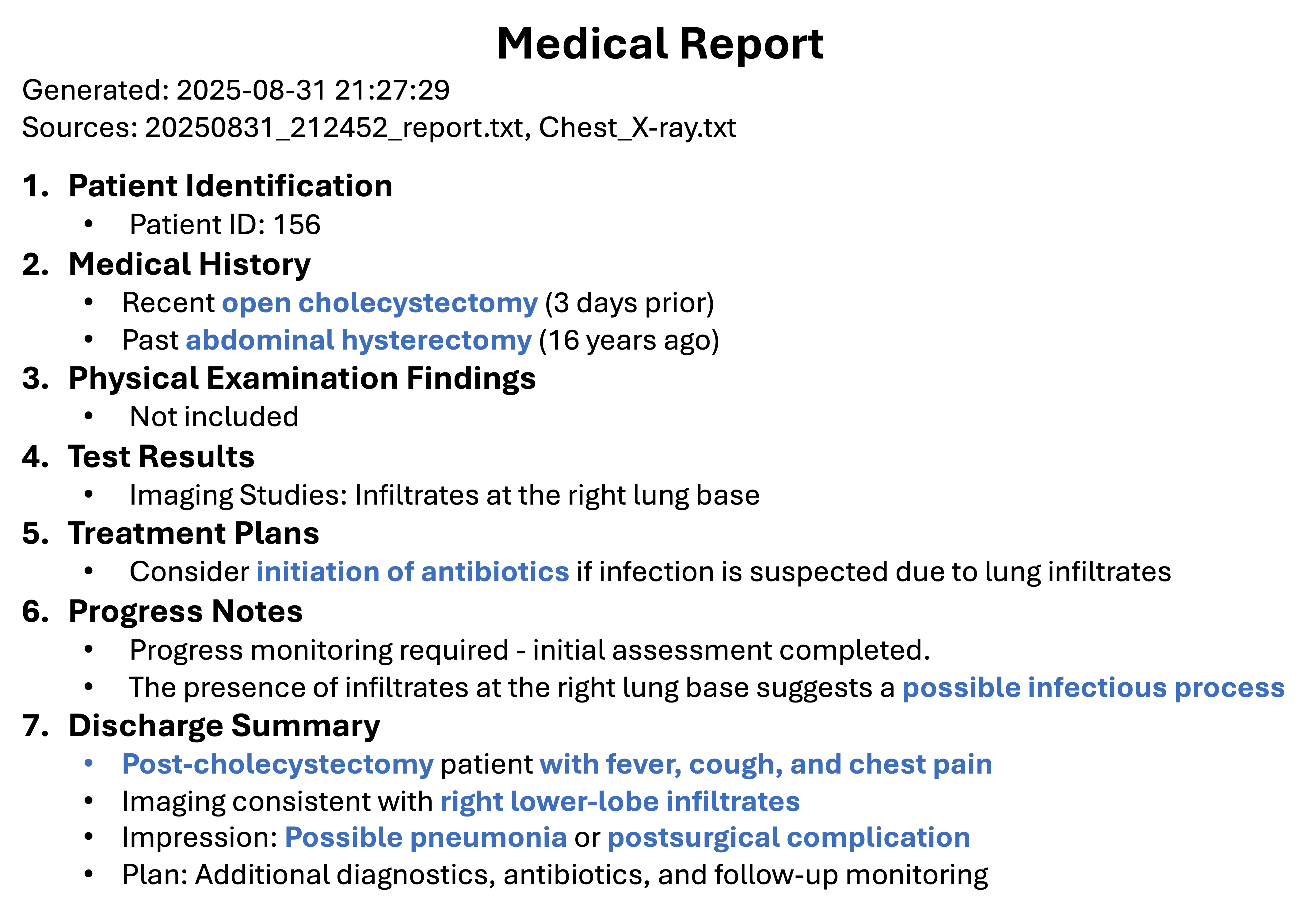}
            \vspace{-10pt}
    \caption{\textbf{A Simple Case Study of the Medical Report.}}
            \label{fig:report_example}
\end{figure}
MedicalOS is designed to generate reports after the initial patient inquiry and following each new examination result. Based on this workflow, the expected average number of reports per patient is 2.53, while the observed average is 2.51. This close alignment demonstrates that MedicalOS reliably follows the intended procedure and consistently produces reports in accordance with the examinations available.

Figure~\ref{fig:report_example} presents a representative case study generated by MedicalOS. The system organized the information into seven standardized sections: Patient Identification, Medical History, Physical Examination Findings, Test Results, Treatment Plans, Progress Notes, and Discharge Summary, while also documenting the data sources used to create the updated report. In this case, the report captured the patient’s recent surgical history (open cholecystectomy), highlighted imaging findings (right lower-lobe infiltrates), and summarized the clinical impression (possible pneumonia or postsurgical complication). It further outlined treatment considerations, including antibiotic initiation and follow-up evaluations.

Together, these results highlight that MedicalOS not only maintains procedural consistency but also delivers concise, clinically relevant summaries. By automatically organizing heterogeneous clinical data into a standardized format, the system improves interpretability and reduces the burden of manual reporting, allowing clinicians to focus on patient care.

\subsubsection{Medication Recommendation Analysis}
\begin{figure}[ht]
            \centering
            \vspace{-10pt}\includegraphics[scale=0.4]{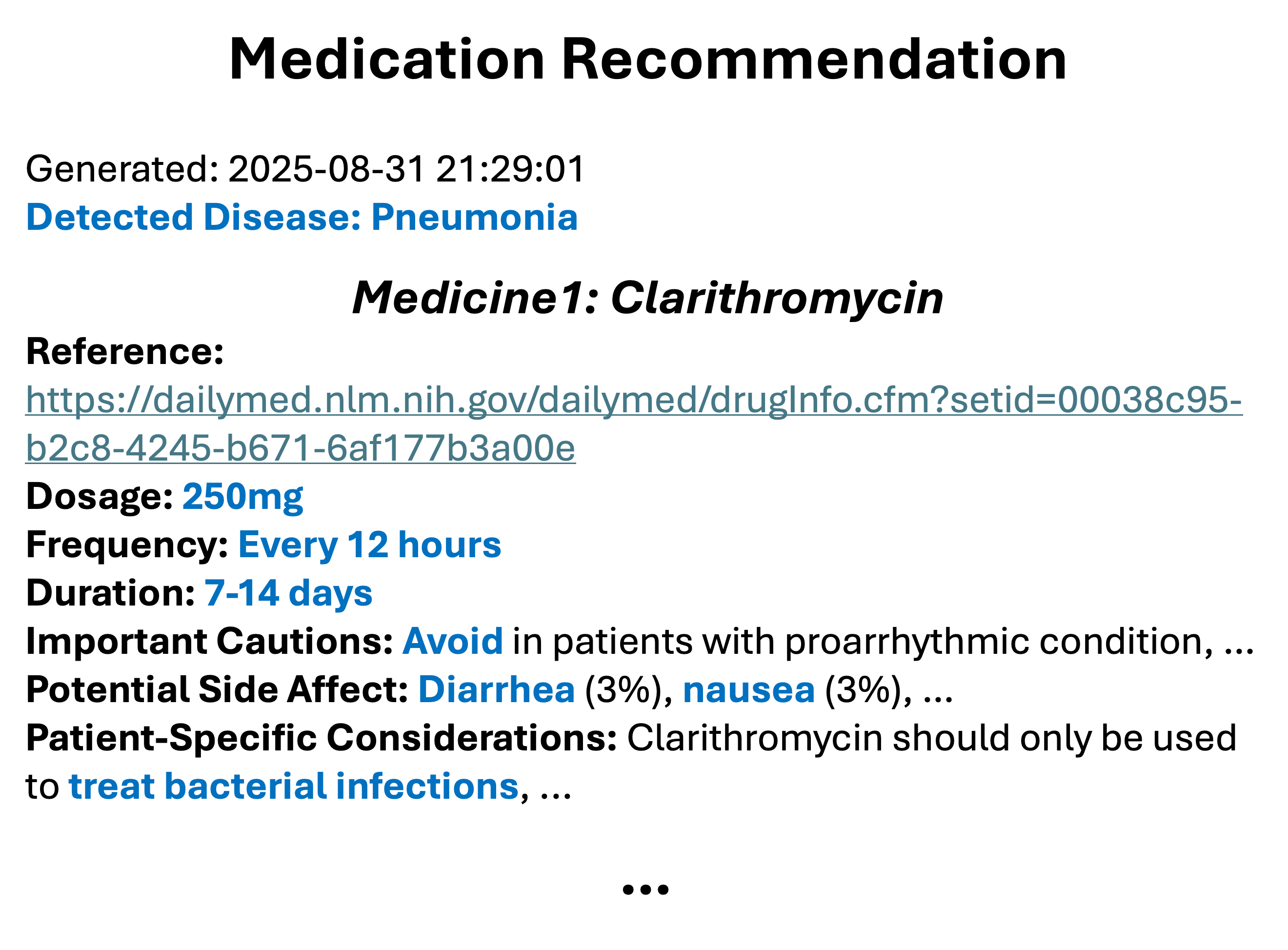}
            \vspace{-20pt}
    \caption{\textbf{A Simple Case Study of the Medication Recommendation.}}
            \label{fig:medication_example}
\end{figure}
MedicalOS is instructed to generate three medication recommendations for each patient at discharge. In practice, MedicalOS closely followed this design: among all cases, 202 reports included three medications, 9 included two, 1 included one, and 2 failed to generate a recommendation. This distribution demonstrates that MedicalOS reliably adheres to its intended prescription protocol, with only rare deviations. The example in Figure~\ref{fig:medication_example} illustrates this process, where the system not only suggested clarithromycin for pneumonia but also provided dosage, frequency, treatment duration, cautions, side effects, patient-specific considerations, and the reference source. This structured output underscores the system’s ability to couple medical reasoning with traceable evidence, supporting both clinical decision-making and transparency.

\section{Conclusion}
In this work, we introduced MedicalOS, a unified agent-based operational system that bridges clinician instructions in natural language with machine-executable actions through a domain-specific abstract layer. MedicalOS enables end-to-end workflow automation, spanning patient inquiry, history retrieval, examination management, report generation, referral, and treatment planning, all while remaining aligned with trusted medical guidelines. Evaluation on 214 patient cases across 22 specialties shows that MedicalOS delivers high diagnostic accuracy and confidence, clinically sound examination requests, and consistent structured reports and medication recommendations. These results highlight its potential to reduce clinician workload, improve transparency, and provide scalable workflow automation in healthcare.



\bibliography{custom}


\end{document}